\documentclass[letterpaper, 10 pt, conference]{ieeeconf}
\IEEEoverridecommandlockouts
\usepackage{times}
\usepackage{epsfig}
\usepackage{graphicx}
\usepackage{amsmath}
\usepackage{amssymb}
\let\labelindent\relax\usepackage[inline]{enumitem}
\usepackage[printonlyused, nolist]{acronym}
\usepackage{bm}
\usepackage{tabularx}
\usepackage{booktabs}
\usepackage{xcolor}
\usepackage{hyperref}
\usepackage[nameinlink]{cleveref}
\usepackage[detect-weight]{siunitx}
\usepackage[activate={true,nocompatibility},final,tracking=true,kerning=true,spacing=true,factor=1100,stretch=20,shrink=20]{microtype}

\usepackage{tikz}
\usepackage{textcomp}
\usepackage{lipsum}

\graphicspath{{images/}{diagrams/}}
\begin{acronym}[UAVs]
\acro{UAV}{Unmanned Aerial Vehicle}
\acrodefplural{UAVs}{Unmanned Aerial Vehicles}
\acro{GPS}{Global Positioning Sensor}
\acro{IMU}{Inertial Measurement Unit}
\acrodefplural{IMUs}{Inertial Measurement Units}
\acro{SLAM}{Simultaneous Localisation and Mapping}
\acro{VO}{Visual Odometry}
\acro{RANSAC}{Random Sample and Consensus}
\acro{EKF}{Extended Kalman Filter}
\acro{ATE}{Absolute Trajectory Error}
\acro{RPE}{Relative Pose Error}
\acro{ROS}{the Robot Operating System}
\acro{RGB}{Red-Green-Blue}
\acro{BA}{Bundle Adjustment}
\acro{LOAM}{Lidar Odometry and Mapping}
\acro{SfM}{Structure from Motion}
\end{acronym}

\newlist{problems}{enumerate}{3}
\setlist[problems,1]{label=\textbf{\arabic*)}, ref=\textbf{(\arabic*)}}
\crefname{problemsi}{}{}

\newcommand{\etal}{\textit{et al.}}
\SetTracking{encoding={*}, shape=sc}{0} 
\title{\LARGE \bf
	A Robust Extrinsic Calibration Framework\\for Vehicles with Unscaled Sensors
}

\author{Celyn Walters, Oscar Mendez, Simon Hadfield, Richard Bowden$^{1}$
	\thanks{$^{1}$C.~Walters, S.~Hadfield, O.~Mendez and R.~Bowden are with CVSSP at the University of Surrey, UK
	{\tt\small c.walters@surrey.ac.uk}}
}

\newcommand\copyrighttext{%
	\footnotesize \textcopyright 2019 IEEE. Personal use of this material is permitted.
	Permission from IEEE must be obtained for all other uses, in any current or future
	media, including reprinting/republishing this material for advertising or promotional
	purposes, creating new collective works, for resale or redistribution to servers or
	lists, or reuse of any copyrighted component of this work in other works.
	DOI: \href{https://doi.org/10.1109/IROS40897.2019.8968244}{10.1109/IROS40897.2019.8968244}
}
\newcommand\copyrightnotice{%
	\begin{tikzpicture}[remember picture,overlay]
	\node[anchor=south,yshift=10pt] at (current page.south) {\fbox{\parbox{\dimexpr\textwidth-\fboxsep-\fboxrule\relax}{\copyrighttext}}};
	\end{tikzpicture}%
}

\begin{document}
\newcommand{\minus}{\scalebox{0.5}[0.75]{\ensuremath{-}}}

\maketitle
\copyrightnotice
\thispagestyle{empty}
\pagestyle{empty}

\begin{abstract}

Accurate extrinsic sensor calibration is essential for both autonomous vehicles and robots.
Traditionally this is an involved process requiring calibration targets, known fiducial markers and is generally performed in a lab.
Moreover, even a small change in the sensor layout requires recalibration.
With the anticipated arrival of consumer autonomous vehicles, there is demand for a system which can do this automatically, after deployment and without specialist human expertise.

To solve these limitations, we propose a flexible framework which can estimate extrinsic parameters without an explicit calibration stage, even for sensors with unknown scale.
Our first contribution builds upon standard hand-eye calibration by jointly recovering scale.
Our second contribution is that our system is made robust to imperfect and degenerate sensor data, by collecting independent sets of poses and automatically selecting those which are most ideal.

We show that our approach's robustness is essential for the target scenario.
Unlike previous approaches, ours runs in real time and constantly estimates the extrinsic transform.
For both an ideal experimental setup and a real use case, comparison against these approaches shows that we outperform the state-of-the-art.
Furthermore, we demonstrate that the recovered scale may be applied to the full trajectory, circumventing the need for scale estimation via sensor fusion.

\end{abstract}

\section{INTRODUCTION}

Autonomous vehicles will be one of the most lifestyle-changing advancements to result from today's Robotics and Computer Vision research.
They have the potential to solve congestion, reduce fatalities and environmental impact.
In order for these agents to operate safely in the real world, they need to interpret their surroundings and robustly localize with a high degree of accuracy.
To this end, modern robotic platforms employ a variety of sensor types, such as stereo/monocular cameras, \acp{IMU}, \acp{GPS} and wheel odometry combined via sensor fusion.

While using multiple sensors provides more robust perception, it is extremely important that the sensors are properly calibrated.
Intrinsic calibration ensures that each individual sensor's output is accurate, and for many commercial sensors, this calibration is performed in the factory.
However, the extrinsic calibration between sensors must be carried out once they are mounted in their final positions.
Generally, this action is avoided by mounting the sensors to a structure with precisely known dimensions.
However, this puts restrictions on the sensor layout and ease of manufacture.
Where this is not possible, the traditional method is to make use of external reference frames~\cite{Zhang2004,Zhao2016}.
Such a procedure is relatively simple to carry out for a test platform, but platforms designed for continuous operation are susceptible to changes in calibration due to vibration and slip in mountings.
This necessitates manual re-calibration, which is inconvenient for larger platforms and vehicles.
Furthermore, a typical consumer will not have access to a calibration environment nor have the expertise to reliably carry out a complicated procedure.

\begin{figure}[t]\centering
	\includegraphics[width=\linewidth]{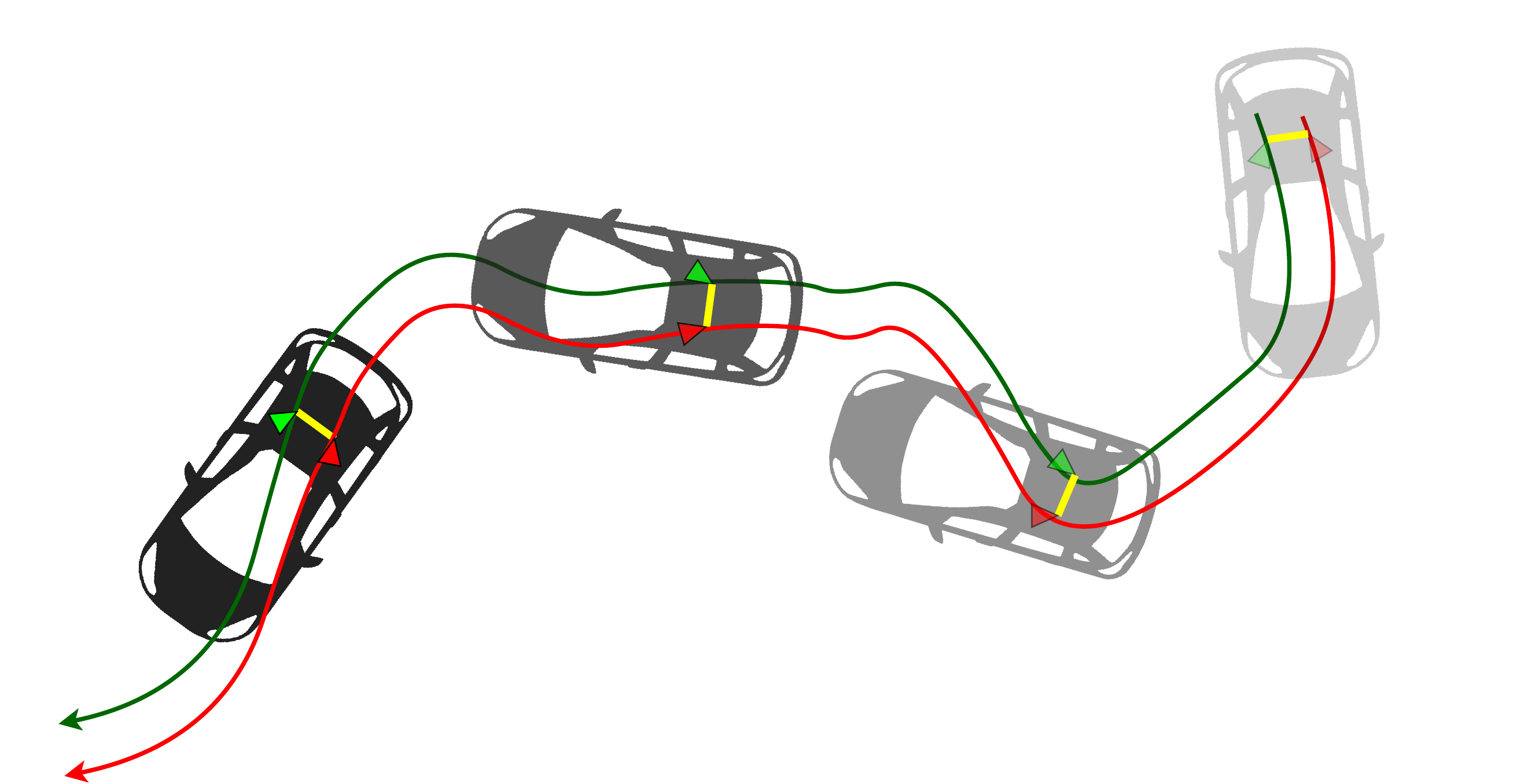}
	\caption{Green and red sensor trajectories are used to recover the fixed extrinsic transform between them, shown as yellow.}
	\label{fig:main}
\end{figure}

An attractive alternative is calibration directly from sensor motion.
Hand-eye calibration techniques~\cite{Chen1991, daniilidis1999hand, Shiu1989} estimate the transform between two rigidly-attached sensors using their poses, as in \cref{fig:main}.
It was originally formulated as a way to recover the transformation between a robotic arm and a mounted camera.
It removes the need for overlapping views and may be broadly applied to visual and nonvisual sensors.
However, there are several limitations to this approach that prevent it from being used as an automatic calibration tool in the field:
\begin{problems}
\item \textbf{Assumes known scale} ---
\label{problem:scale}
A prerequisite for hand-eye calibration is that the measurements are to a consistent scale, usually metric.
This is not the case for monocular cameras, and the common solution is to rely on visible external reference frames or to rescale the sensor using another metric sensor such as \ac{GPS}.

\item \textbf{Corruption by degenerate motion} ---
\label{problem:degenerate}
For a complete description of the relative transformation, hand-eye calibration requires stimulation of all degrees of freedom.
Without appropriate motion, parts of the extrinsic transform will be unobservable.
It is impractical for a consumer to be required to carry out calibration manoeuvres, but the necessary movements may occur during prolonged natural use.
The truly unobservable parts, such as impossible vehicle motion, are by nature not important for sensor fusion-based odometry.

\item \textbf{Vulnerable to drift} ---
\label{problem:drift}
Individual sensors exhibit compounded error in the form of drift.
While additional data can increase the accuracy of the extrinsic calculation,
acquiring data over a longer period exacerbates drift.
\end{problems}

In this paper, we present a single flexible solution which can
overcome all of these issues.
\Cref{problem:scale} is addressed by simultaneously solving for the extrinsic transform and its corresponding scale by optimizing a similarity matrix.
This allows monocular sensors to be calibrated in the same way as any metric sensor.
This means that algorithms which operate on unscaled sensors such as \ac{VO} may be combined with other data sources in a metric sensor fusion framework.
To address \cref{problem:degenerate}, the calibration runs in real-time without explicit user intervention.
The agent does not need to perform specialized movements, as the framework updates each parameter when it is observable, providing robustness against periods of insufficient movement or lack of visual features.
For \cref{problem:drift}, our framework mitigates the effects of drift caused by poor visibility and noise, which commonly interfere with \ac{VO} \& \ac{IMU} processing.
We exploit the fact that data is often locally accurate, and combine several overlapping sections of the trajectory which contain less drift.
In addition, the calibration discards the sections containing discontinuities
that may be generated, for example, by \ac{GPS} corrections, or by a loop closure in a visual odometry system.

Since direct measurements are not used, the calibration is indifferent to the source of data, allowing remarkable flexibility.
For instance, a \ac{VO}/\ac{SLAM} algorithm may be freely chosen.

This paper demonstrates how our framework achieves accurate extrinsic measurements even when presented with imperfect input data.
\Cref{sec:literature} surveys other extrinsic calibration approaches and the foundations for this work.
\Cref{sec:handeye} describes how preliminary estimates for transformation and scale are generated.
In \cref{sec:robust}, the estimates are selected and processed to account for measurement inaccuracies and degenerate motion.
Finally, the performance is evaluated in \cref{sec:zeds,sec:baseline,sec:turtlebot,sec:kitti} against competing approaches.

\section{RELATED WORK}
\label{sec:literature}

A simple approach to general extrinsic calibration is described by Zhao \etal~\cite{Zhao2016}, in which markers attached to camera housings are simultaneously viewed by an external `support' camera.
This way, even non-overlapping cameras can be simply calibrated without degenerate cases.
Although flexible, the method depends on exact measurements between each sensor and its attached marker which may be difficult to obtain in practice.

Traditional approaches to extrinsic calibration exploit sensor-specific capabilities to maximize accuracy.
Zhang and Pless use a calibration grid to retrieve the extrinsic transform between a camera and a 2D laser range finder~\cite{Zhang2004}.
Velas \etal~replace the calibration grid with a novel 3D marker~\cite{Velas2014}.
For intrinsic and extrinsic calibration of \acp{IMU} and RGB cameras, the Kalibr calibration toolbox by Furgale \etal~may be used~\cite{Furgale2013, Rehder2016}.
All of these rely on a dedicated calibration procedure, which may not be practical for a large vehicle that cannot undergo certain motions.
A more automated method is used in our framework, in which the calibration data is obtained automatically during normal operation.

An approach which is less dependent on specific visible cues is to instead use natural features.
Ling and Shen achieve this for finding the offset between each sensor in a stereo camera~\cite{Ling2016}.
For monocular cameras, Ataer-Cansizoglu \etal~exploit a previously generated \ac{SLAM} model, where 2D-3D correspondences are formed from the images to find the global camera positions~\cite{Ataer-Cansizoglu2015}.
More closely related to our work is the approach of Maye \etal~, in which the self-calibration is online and robust to small motion~\cite{Maye2013}.
Ours differs as it does not directly use landmarks nor assume these are available.

\begin{figure*}\centering
	\includegraphics[width=\linewidth]{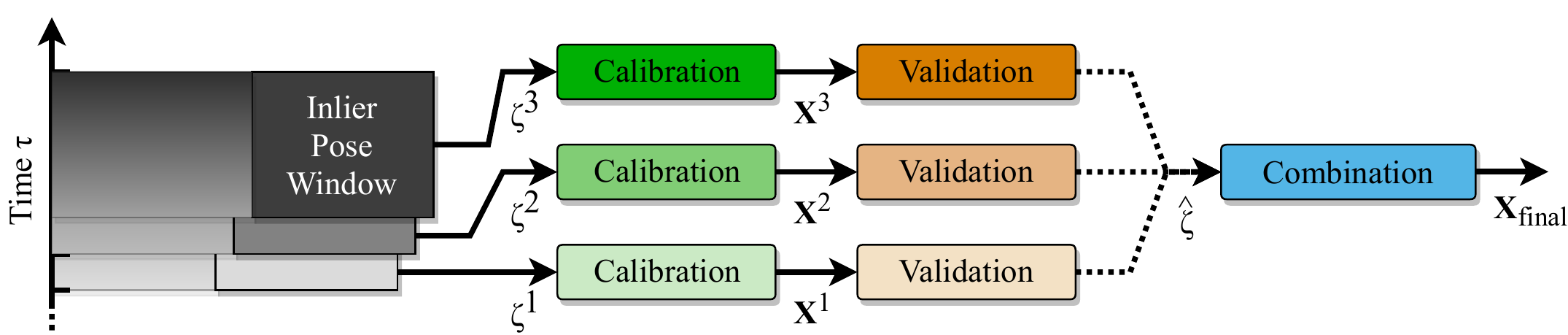}
	\caption{Systems diagram of calibration framework.}
	\label{fig:system}
\end{figure*}

Most algorithms which directly process visual features are restricted to those which use known geometry or where there are overlapping views.
Our approach, which builds on hand-eye calibration, requires only relative sensor pose information.
Solutions for hand-eye calibration were published as early as 1989 by Shiu and Ahmad~\cite{Shiu1989}.
Early methods estimate the rotation and translation independently.
Traditionally, rigid transformation is decomposed in this way to follow the rotation-then-translation formalization.
This leads to a common perception that their relationship is separate.
In~\cite{Chen1991}, Chen argues that decoupling the two adversely affects the generality and efficacy of the algorithm.
They approach the problem with screw motion to unify them as a single parameter.
Daniilidis also jointly parameterizes rotation and translation, this time using dual quaternions~\cite{daniilidis1999hand}.

The problem of scale is often overlooked, as it can usually be obtained through a separate process.
Although Schmidt \etal~extend~\cite{daniilidis1999hand} by encoding scale as the norm of the dual quaternion~\cite{Schmidt2005}, the rotation, scale, and translation are obtained separately in evaluation.
The relative error also seems high and the scale error is not provided.
Heller \etal~use second-order cone programming to recover the scaled translation component.
Again, the rotation is separately calculated, and convergence time is in the order of minutes.
In contrast to the above approaches, we opt to estimate both the extrinsic transform and corresponding scale in the same operation in real-time.

Vehicles with restricted motion modes cannot stimulate all the necessary degrees of freedom required for a full hand-eye calibration.
For example, for wheeled vehicles which can only travel with planar motion and evidence of rotation is limited to a single axis, a sensor's height is unobservable.
Ruland \etal~and Heng \etal~both apply hand-eye calibration for cameras in a vehicular context, although they ignore the estimation of relative heights, or rely on a separate technique~\cite{Ruland2010,Heng2013}.
Heng \etal~utilizes the dual-quaternion approach from~\cite{daniilidis1999hand} and makes use of the visual features between views as a refinement in an offline process.
In our work, all parameters come from the same optimization and are updated if and when they are recoverable.

Zhu \etal's process extends hand-eye calibration by exploiting generic planes using \ac{SfM}~\cite{Zhu2016}.
This allows calibration of non-overlapping views from multiple cameras.
Since our method does not assume visual information is available, cameras may be treated as black boxes which provide unscaled pose information.
This is advantageous in the case of cameras which process \ac{VO}/\ac{SLAM} algorithms on-board, and allows simpler data connections for a more modular setup.

\section{METHODOLOGY}
\label{sec:method}

State-of-the-art approaches cater for specific use-cases, and often rely on particular sensors.
Instead, we propose an extrinsic calibration framework, laid out in \cref{fig:system}, which allows fully modular configuration of sensors and their pose-estimation algorithms.
Its operation may be summarized as follows:
First, sets of time-aligned poses from two arbitrary sensors are collected (black blocks, \cref{fig:system}).
Estimates for the extrinsic calibration between the sensors are obtained using each pose set (green blocks), and described in \cref{sec:handeye}.
The sets are filtered based on the type of motion and the quality of the estimate determined (orange blocks), and is described in detail in \cref{sec:robust}.
Finally, the inlier estimates are continuously combined (blue block) for a stable calibration result which is refined over time.

\subsection{Extrinsic Calibration}
\label{sec:handeye}

\begin{figure}[h]\centering
	\includegraphics[width=\linewidth]{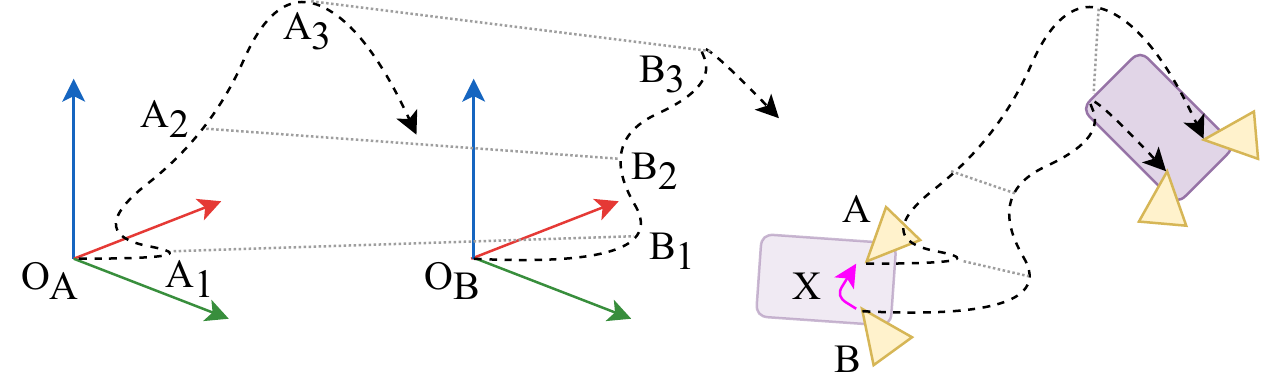}
	\caption{Concept of hand-eye calibration for two cameras \(A\) and \(B\). Poses \(\bm{A}_{1-3}\) and \(\bm{B}_{1-3}\) are recorded in an unknown frame relative to their original position \(\bm{O}_A\) and \(\bm{O}_B\). When correctly placed, their trajectories have constant relative transformation \(X\).}
	\label{fig:handeye}
\end{figure}

There are multiple formulations for hand-eye calibration but the most widely-used solves the following equation:
\begin{equation}
	\label{eq:axxb}
	\bm{A} \bm{X} = \bm{X} \bm{B}
	,
\end{equation}
where \(\bm{A}\), \(\bm{B}\) and \(\bm{X}\) are homogeneous transformations in the form
\begin{align}
	\label{eq:matrix}
	\bm{A},\bm{B} =
	\begin{bmatrix}
		\bm{R} & \bm{t}\\
		0^T & 1
	\end{bmatrix}
	,&&
	\bm{X} = s
	\begin{bmatrix}
		\bm{R} & \bm{t}\\
		0^T & 1/s
	\end{bmatrix}
	.
\end{align}
\(\bm{A}\) and \(\bm{B}\) represent the pose for the first and second sensors respectively.
By rearranging \cref{eq:axxb} into \(\bm{A} = \bm{X} \bm{B} \bm{X}^{\minus1}\), it can be seen that the pose from sensor \(\bm{B}\) transformed by the correct transformation \(\bm{X}\) will be identical to the pose from sensor \(\bm{A}\).
Assuming that \(\bm{A}\) and \(\bm{B}\) are rigidly attached, the same transform \(\bm{X}\) will solve for a series of \(\bm{A}\) and \(\bm{B}\) poses.
Therefore, \(\bm{X}\) represents the constant unknown transformation that maps the reference frame of \(\bm{A}\) to that of \(\bm{B}\).
This concept is shown in \cref{fig:handeye}.

For calibration, a series of time-synchronized poses \(\zeta_A\) and \(\zeta_B\) are recorded for each sensor relative to \(\bm{O}_A\) and \(\bm{O}_B\) respectively, where
\begin{equation}
	\label{eq:poses}
	\begin{array}{l}
		\zeta_A = \left\{\bm{A}_0, \bm{A}_1 \cdots \bm{A}_\tau\right\}\\
		\zeta_B = \left\{\bm{B}_0, \bm{B}_1 \cdots \bm{B}_\tau\right\}
		,
	\end{array}
\end{equation}
and \(\bm{O}\) is defined as the initial position of each set of poses \(\zeta\), such that
\begin{equation}
	\label{eq:eachorigin}
	\begin{array}{l}
		\bm{O}_A \bm{A}^{\minus1}_0 = \bm{I}
		,
	\end{array}
\end{equation}
where \(\bm{I}\) is the identity transform matrix.
The reason that this is necessary is that the sets of poses \(\zeta_A\) and \(\zeta_B\) are independent and are defined in their own arbitrary reference frame.
It is not important what sensor this data originates from, which gives rise to the flexibility of the approach.
Poses from cameras may be obtained through any \ac{VO} or \ac{SLAM} algorithm, which are abundantly available~\cite{Forster2017, Mur-Artal2017, Zhang2015a}.
For a given autonomous/robotic platform, it is likely that such an algorithm already needs to be run for sensor fusion and mapping purposes.
This being the case, there is no additional computational cost in terms of obtaining these poses.
Similarly, there are odometry solutions available for lidar scanners, and poses may also be obtained from \acp{IMU}, \ac{GPS} or wheel odometry.

Hand-eye calibration is performed using \(\zeta_A\) and \(\zeta_B\), resulting in an estimate for the extrinsic transformation.
To optimize for scale in the same calibration step, it is necessary to be able to incorporate a scale parameter into each input measurement \(\bm{A}_\tau\) and \(\bm{B}_\tau\).
For this reason, we use a similarity matrix.
This necessitates the use of a nonlinear optimization, using the Frobenius norm in the cost function
\begin{equation}
	\label{eq:costfunction}
	h(\bm{X} | \zeta_A, \zeta_B) = \sum_{\tau}^{ } \left\Vert \bm{A}_\tau \bm{X} - \bm{X}\bm{B}_\tau \right\Vert
	.
\end{equation}
\(\bm{A}_\tau\) and \(\bm{B}_\tau\) are the synchronized poses of the first and second sensor respectively, and \(\bm{X}\) is the extrinsic estimate.
The problem is minimized using the Levenberg-Marquardt algorithm initialised with identity transform or, if one exists, the previously known calibration.
If both sensors are metric, the result of scale parameter \(s\) from \cref{eq:matrix} is usually near \(1.0\).
Fixing the value reduces the degrees of freedom and allows for higher accuracy.

\subsection{Robust Estimation Pipeline}
\label{sec:robust}

In a real-world scenario, data is subject to noise and drift.
The minimization does not make adjustments to the input poses as this would require sensor-specific techniques such as \ac{BA} for cameras and bias estimation for \acp{IMU}.
Optimizing over longer periods makes it more likely that all 6 degrees of freedom are exercised, enabling a fully observable solution.
On the other hand, this also incorporates more error due to drift.

The following describes our methods which bring robustness to our scaling hand-eye calibration, allowing it to run in a deployed system without supervision.
To allow isolation of recorded data which is subject to drift or discontinuity, intermediate calibrations are repeatedly carried out over a sliding window, shown by the vertical axis in \cref{fig:system}.
This gives frequent estimates \(\bm{X}^i\) of the extrinsic transformation,
\begin{equation}
	\label{eq:intermediate}
	\bm{X}^i = \arg \min_{\bm{X}} \left(h(\bm{X} | \zeta^i_A, \zeta^i_B)\right)
	,
\end{equation}
where \(h\) is our scaling hand-eye calibration and \(i\) is the iteration.

For a complete solution including all axes of translation, the input data needs to exercise rotational motion in each of the X, Y, and Z axes.
For example, in a movement with no rotation, all points rigidly attached to an agent produce identical trajectories relative to their starting positions.
In this case, there are infinitely many values of \(X\) which have an identical minimizer cost.
Many vehicles do not rotate freely and spend the majority of the time moving in ways which do not allow a full extrinsic calibration.
By selecting windows \(\zeta^i\) based on their eigenvalues, we can ensure that they contain enough motion.
For nonholonomic vehicles, the ratio between eigenvalues computed over position indicates rotation.
This allows our system to automatically determine which parts of the extrinsic transformation can reliably be estimated and disable calibration when stationary.

Some windows inevitably contain errors or degenerate motion, leading to incorrect extrinsic estimates.
Early rejection is applied to remove the most severe failures by thresholding their respective optimizer costs.
When problematic input results in a low cost, the produced translation or orientation is an obvious outlier and is removed using RANSAC.
Overlapping windows allows a greater number of calibrations to be obtained, giving stability to RANSAC.
Although pose pairs may be incorporated into multiple overlapping windows, each is relative to a different reference frame and therefore estimates may be treated independently.
The mean translation and rotation over all inlier windows consolidates and refines the accumulated evidence over longer periods.

In contrast to other calibration approaches, ours can take advantage of existing knowledge.
It is often possible to physically measure the distance between sensors.
The function in \cref{eq:intermediate} may be extended with a regularizer term,
\begin{equation}
	\label{eq:regulariser}
	\bm{X}_{\mathrm{final}} = \arg \min_{\bm{X}} \left(h(\bm{X} | \hat{\zeta_A}, \hat{\zeta_B}) + \alpha\omega\right)
	,
\end{equation}
where \(\alpha\) is a weight and \(\omega\) is the difference between the Euclidean distance of the extrinsic estimate and the scalar measurement.
This results in a significantly tighter estimate cluster and may even overcome unobservable axes.
Since the transformation is not expected to dramatically change, the previous calibration can also be used as a prior guess.
Even so, accumulation of evidence allows the system to recover from situations such as sensor mountings shifting during operation.

In summary, this section introduced the concept of hand-eye calibration and how it applies generally to the task of recovering relative extrinsic transforms.
It also highlighted several limitations which reduce the effectiveness of traditional hand-eye calibration in real scenarios.
These include calibrating with flawed input data, and compromises between gathering sufficient motion and minimizing drift.
Methods by which our framework addresses these limitations were explained.
The following section shows that these solutions are effective in comparison to previously published work.

\section{RESULTS}

This section demonstrates that the capabilities of our framework allow it to achieve competitive accuracy with other hand-eye calibration methods when exposed to ideal movement and known scale.
In the absence of these, we show that the approach significantly outperforms other techniques.
The performance of the framework is evaluated in three scenarios:
firstly, for a hand-held platform to provide nondegenerate movement;
secondly, for a ground-based robotic platform using monocular vision restricted to planar motion;
and finally, using a well-known benchmark with multiple sensor types to test performance in real-world conditions.

\subsection{Effect of Baseline}
\label{sec:baseline}

\begin{figure}[h]\centering
	\includegraphics[width=\linewidth]{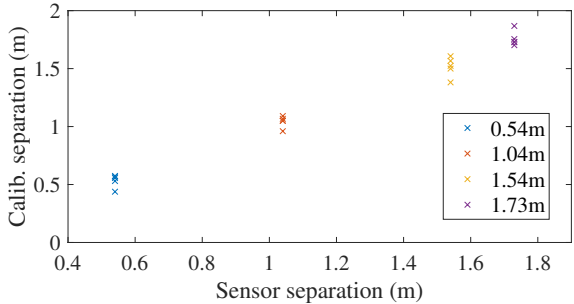}
	\caption{Position of calibration results at different baselines.}
	\label{fig:sep}
\end{figure}

A test was performed to examine the effect of baseline length between sensors has on hand-eye calibration position accuracy.
Two matching stereo cameras were fixed to a rigid bar at different spacings, with translation constrained to a single axis and no rotation.
Camera motion was extracted using stereo ORB-SLAM2~\cite{Mur-Artal2017} in a feature-rich environment with smooth motions and each test was repeated five times.
The even spread around the correct distance at each separation in \cref{fig:sep} shows that wider baselines do not significantly affect the accuracy, and it is not valid to evaluate calibration error as a percentage of the ground truth.
Hand-eye calibration has stronger dependence on the accuracy of input poses.

\subsection{Twin Stereo Cameras}
\label{sec:zeds}

\begin{figure}[h]\centering
	\includegraphics[width=0.7\linewidth,clip,trim=0 1cm 0 0.6cm]{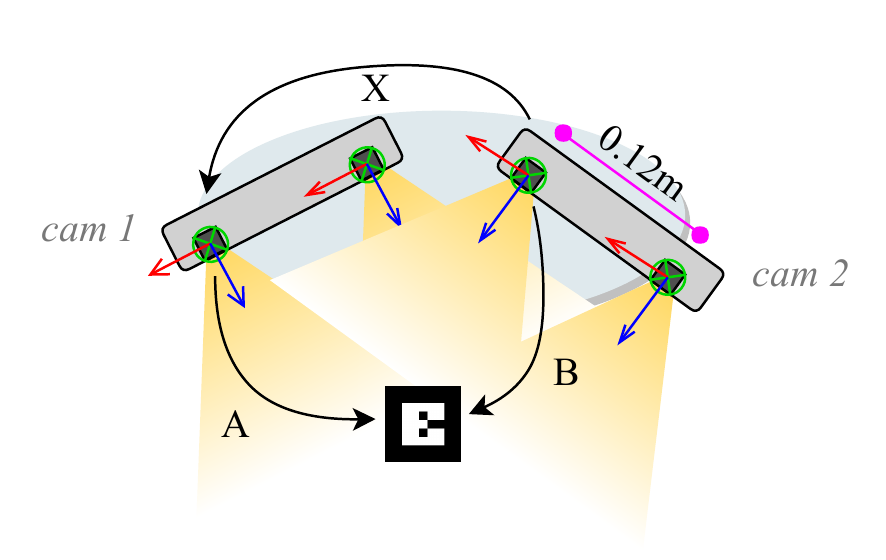}
	\caption{Overview of stereo camera layout.
	Relative 3D rotation is arbitrary.}
	\label{fig:zeds}
\end{figure}

To evaluate real-world performance in the best case, the data needs to be in metric scale and perform motion which exercises all six degrees of freedom to make sure all parameters are observable.
The input data should be as error-free as possible.
This demands slow, smooth movement with well-lit, static surroundings with plenty of features.
To fulfil these needs, a platform was assembled for dataset capture.
Two matching stereo cameras (Stereolabs ZED cameras) were mounted to a rigid base, shown in \cref{fig:zeds}.

The ground truth transforms \(A\) and \(B\) between each camera and the ARTag were obtained using a tracking library~\cite{Rohs2008}.
These were used to calculate the ground truth relative transform \(X\) between cameras.
The overlapping views and ARTag are only for ground truth measurement;
they do not benefit the calibration.
To isolate \ac{VO} error, poses from \textit{cam 1} were obtained using
stereo \ac{VO}~\cite{zed}, and poses from \textit{cam 2} were the same transformed by the ground truth.
The base was moved to excite all degrees of freedom over 90 seconds.
Predictably, all hand-eye calibration approaches give very close to zero error.
However, using independent \ac{VO} for \textit{cam 2}'s poses instead shows a significant loss of accuracy and the necessity for robustness.

\begin{table}
\centering
\footnotesize
\caption{Translation and rotation calibration error, and evaluation of resulting trajectories using the calibration}
\label{tab:zeds}
\begin{tabular}{cccccc}
	\toprule
	Calibration & T (m) & R (\(^\circ\)) & ATE (m) & \multicolumn{2}{c}{RPE (m, \(^\circ\))}\\
	\midrule
	Schmidt~\cite{daniilidis1999hand} (full) & 0.2118 & 2.68 & 0.188 & 0.075 & 2.07\\
	Ours (full) & 0.0620 & 3.21 & 0.159 & 0.049 & 2.15\\
	Schmidt (robust) & 0.0270 & 1.57 & 0.162 & 0.039 & 1.71\\
	Ours (robust) & \textbf{0.0105} & 1.49 & \textbf{0.147} & 0.039 & \textbf{1.64}\\
	Ours (scale) & 0.0276 & \textbf{1.41} & 0.160 & \textbf{0.038} & 1.65\\
	\bottomrule
\end{tabular}
\end{table}

\Cref{tab:zeds} lists the distance and shortest rotation from the ground truth for the following methods:
\begin{enumerate*}
	\item The dual quaternion method of~\cite{daniilidis1999hand} over the entire trajectory,
	\item our Frobenius norm parameterization as shown in \cref{eq:costfunction},
	\item \cite{daniilidis1999hand} using our robust pipeline,
	\item our parameterization using the robust pipeline,
	and \item our robust method with scale optimization.
\end{enumerate*}
The root mean squared error of \ac{ATE} and \ac{RPE}~\cite{Sturm2012} is calculated using the trajectory from \textit{cam 1} and \textit{cam 2} after transforming by the inverse extrinsic transform result.
Drift and small loop closures (which are widespread in robotic vehicles) corrupt the global trajectories despite efforts to minimise them.
As with other hand-eye calibration techniques, the dual quaternion approach has been shown to be highly accurate~\cite{daniilidis1999hand,Schmidt2005,Heng2013}, but only in situations with significantly less measurement noise.
\Cref{tab:zeds} shows our approach is more suitable in this area, and also that redundantly estimating scale does not detract from the overall accuracy.

\subsection{Turtlebot}
\label{sec:turtlebot}

A second dataset was required to demonstrate our approach's benefits to automatic scale estimation and robust self-calibration in the context of robotics.
We used the dataset from Mendez \etal~\cite{Mendez2017}, consisting of a ground-based robotic platform, Turtlebot, travelling with periods of rotation and of linear motion.
A forward-facing Microsoft Kinect depth sensor was mounted with a known relative transform, to be retrieved by our calibration.
The ground truth trajectory was generated using Labb{\'{e}} and Michaud's RGB-D \ac{SLAM} algorithm~\cite{Labbe2014}, and registered to the floorplan.
With yaw rotation only, the relative sensor heights are unobservable.
However, as the sensor outputs are relative to the plane of motion, the height difference is inconsequential for applications such as sensor fusion.
The pose sources were the wheel odometry fused with gyroscope, and monocular \ac{VO} (SVO 2.0~\cite{Forster2017}) using the Kinect \ac{RGB} image.
Our calibration does not use the depth information.
Monocular \ac{VO} typically derives its scale from the mean depth of tracked points and therefore depends on the dynamic scene view.
Without incorporating global optimization such as bundle adjustment, the \ac{VO} exhibits scale variation over a sequence and every time it reinitializes.
In contrast, our calibration derives scale from relative transformation between sensor positions which is static.

\begin{figure}[h]\centering
	\includegraphics[width=\linewidth]{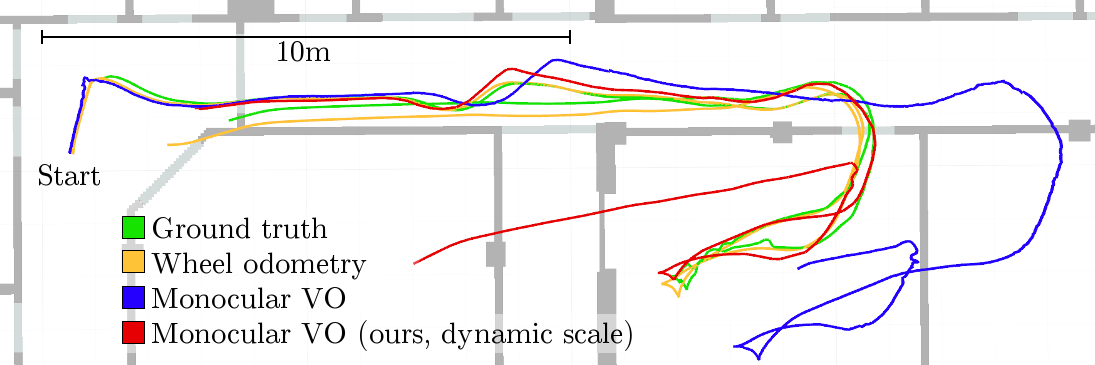}
	\caption{Turtlebot path from different sources overlaid onto the floorplan.
	The dynamically scaled version of monocular \ac{VO} brings it much closer to the ground truth.}
	\label{fig:trajectories}
\end{figure}

A qualitative evaluation of the scale recovery can be made with \cref{fig:trajectories}.
The initial scale of the \ac{VO} (blue) has been manually matched to metric.
Its separate trajectory on the right of the figure is caused by positive scale drift,
and loss of visual tracking causes the sharp corner near the end.
By applying our continuous scale estimation (red), the path closely resembles the ground truth (green) and makes it through the doorway.
The final section is somewhat recovered, though it is heavily dependent on the tracking quality.
Global drift is not an issue when being fused with an absolute positional reference, such as \ac{GPS}.

\begin{figure}[h]\centering
	\includegraphics[width=\linewidth]{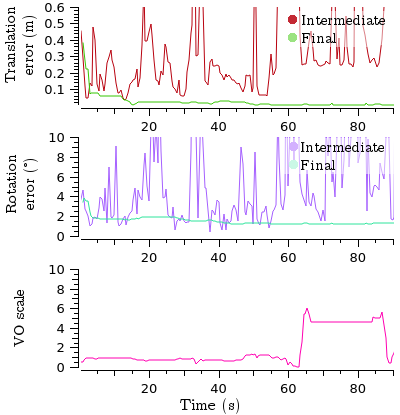}
	\caption{Error, scale for unfiltered estimates and updating calibration.}\label{fig:comparison}
\end{figure}

The unfiltered intermediate and combined calibrations are plotted over time in \cref{fig:comparison}.
The noise in the unfiltered estimates clearly shows the need for intelligent data selection, and the diminishing error of the final transform is a result of the accumulation of evidence.
The scale increase at 60 seconds corresponds with poor \ac{VO} on the final turn, seen in \cref{fig:trajectories}.

The final calibration errors can be found in \cref{tab:turtlebot}.
`Schmidt (scaled)' used a 40 second section with with minimal scale drift and was manually scaled before calibrating.
Unsurprisingly, incorrectly scaled monocular values cause standard hand-eye calibration to fail.
The final row uses the regularizer with weight \(\alpha=0.1\) to work in SE3 space (includes the `unobservable' vertical component).
It also yields slightly better rotation estimates but requires an initial guess.
Here our framework overcomes both the scale issue and \ac{VO} shortcomings without directly using sensor data.

\begin{table}
\centering
\footnotesize
\caption{Translation error when calibrating monocular \ac{VO} to wheel odometry + gyro, for two different sequence lengths}
\label{tab:turtlebot}
\begin{tabular}{cccccc}
	\toprule
	Calibration & T (m) & R (\(^\circ\)) & ATE (m) & \multicolumn{2}{c}{RPE (m, \(^\circ\))}\\
	\midrule
	Schmidt~\cite{daniilidis1999hand} & 0.3126 & 1.28 & 0.3175 & 0.6535 & 27.24\\
	Schmidt (scaled) & 0.1116 & 1.73 & 0.0987 & 0.6507 & 25.91\\
	Ours & \textbf{0.0139} & 1.31 & \textbf{0.0755} & \textbf{0.6391} & 26.95\\
	Ours (6Dof) & 0.0239 & \textbf{1.04} & 0.0924 & 0.6396 & \textbf{24.34}\\
	\bottomrule
\end{tabular}
\end{table}

\subsection{Dataset Benchmarks}
\label{sec:kitti}

As the focus of our work is robustness during normal operation in real scenarios, a third experiment was performed to gauge performance on representative data: The well-established KITTI dataset~\cite{Geiger2013IJRR}.
`Odometry' sequences contain stereo images from two cameras (grayscale and \ac{RGB}), 3D Velodyne point cloud and fused \ac{GPS}/\ac{IMU}.
The sensor extrinsics were accurately calibrated by the dataset authors on each day of recording.

We obtained the camera trajectories using stereo ORB-SLAM2~\cite{Mur-Artal2017}, and
the lidar poses were generated using an open-source version of Zhang and Singh's \ac{LOAM}~\cite{Zhang2015a}.
The \ac{SLAM} maps are not seen by our calibration.
The vertical translation component is evaluated to be unobservable due to rotations only on the ground plane.
As with the Turtlebot example however, the vertical offsets would not affect their reported motion and therefore are not required in sensor fusion.
Introduction of a weak regularizer as outlined in \cref{eq:regulariser} allows recovery of the vertical displacement, despite being defined only by a scalar distance.

\begin{figure}\centering
	\includegraphics[width=\linewidth]{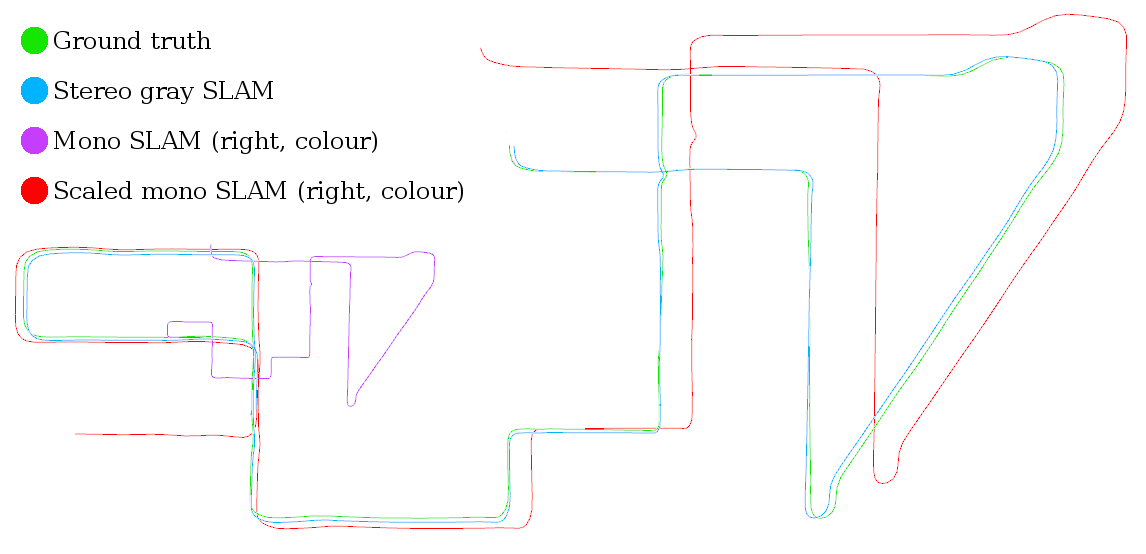}
	\caption{\ac{SLAM} trajectories from odometry sequence 8.
	The scaled version of monocular \ac{SLAM} locally matches stereo \ac{SLAM} and the ground truth.}
	\label{fig:kitti_path}
\end{figure}

\begin{table}
\centering
\footnotesize
\caption{Calibration error between different sensors with respect to the ground truth for KITTI odometry sequence 8}
\label{tab:kitti}
\begin{tabular}{ccc}
	\toprule
	\multicolumn{3}{c}{\textit{Poses in metric scale}}\\
	Sensors calibrated & T (m) & R (\(^\circ\))\\
	\midrule
	\ac{GPS}/\ac{IMU} \(\rightarrow\) lidar & 0.0311 & 0.23\\
	Stereo \ac{SLAM}, gray \(\rightarrow\) color & 0.0190 & 0.31 \\
	Lidar \(\rightarrow\) stereo \ac{SLAM} color & 0.0277 & 0.13\\
	Lidar \(\rightarrow\) stereo \ac{SLAM} color (3DoF) & 0.0877 & 0.77\\
	\midrule
	\multicolumn{3}{c}{\textit{Solving for scale}}\\
	\midrule
	\ac{GPS}/\ac{IMU} \(\rightarrow\) mono color & 0.0518 & 0.19 \\
	Stereo \ac{SLAM} gray \(\rightarrow\) mono color & 0.0629 & 0.25 \\
	Stereo \ac{SLAM} gray \(\rightarrow\) mono color (3DoF) & 0.0633 & 0.10 \\
	\bottomrule
\end{tabular}
\end{table}

\Cref{tab:kitti} shows the results of our calibration for a selection of sensor pairs (with and without scale estimation) compared with the dataset calibrations.
Again, the rows noted `3DoF' make use of the regularizer (\(\alpha=0.1\)) and include the usually unrecoverable vertical component.
Scale in monocular \ac{SLAM} drifts over time, but our calibration can correct for it when the motion allows.
A visualisation of the scale recovery is shown in \cref{fig:kitti_path} (purple to red).
The misalignment is partially a result of comparing separate non-deterministic \ac{SLAM} instances.
Scale estimates are based on accumulated data and may also be slightly delayed, but it shows subsections are locally accurate.

\addtolength{\textheight}{-7.3cm}
\section{CONCLUSION}

In conclusion, this work has demonstrated that robust hand-eye calibration is both convenient and can be more effective than other extrinsic retrieval techniques.
We demonstrated how providing robustness to a constrained technique allows it to gather evidence over time and overcome sections of degenerate or noisy data.
This is in addition to metric scale recovery which enables fusion of monocular sensors.
It is dependent on the quality of the sensor odometry, so best results come from state-of-the-art algorithms in feature-rich environments.

Commercial robots and autonomous vehicles have complex sensor configurations and must be made as a single unit if they are to be factory calibrated.
A tailored calibration procedure can be expected to yield a higher degree of accuracy.
However, our approach displays the potential for exceptional flexibility, where existing vehicles of different brands and models could be retro-fitted with sensors without the need for a specialized procedure.

{
	\small
	\bibliographystyle{IEEEtran.bst}
	\bibliography{bibliography}
}

\end{document}